\documentclass[11pt, conference, letterpaper]{IEEEtran}
\IEEEoverridecommandlockouts
\usepackage[utf8]{inputenc}
\usepackage[T1]{fontenc}
\usepackage{cite}
\usepackage[numbers,sort&compress]{natbib}
\usepackage{amsmath,amssymb}
\usepackage{algorithmic}
\usepackage{graphicx}
\usepackage{textcomp}
\usepackage{xcolor}
\DeclareUnicodeCharacter{FF1A}{:}
\usepackage{url}
\usepackage{booktabs}
\usepackage{nicefrac}
\usepackage{microtype}
\usepackage{tabularx}
\usepackage{float}
\usepackage[skip=4pt,font=small]{caption}
\usepackage{etoolbox}
\makeatletter
\newcommand{\conftopline}{}
\newcommand{\setconftopline}[1]{\renewcommand{\conftopline}{#1}}
\pretocmd{\@maketitle}{%
  \begingroup
  \centering\small\conftopline\par\vspace{0.6em}%
  \endgroup
}{}{}
\makeatother

\def\BibTeX{{\rm B\kern-.05em{\sc i\kern-.025em b}\kern-.08em T\kern-.1667em\lower.7ex\hbox{E}\kern-.125emX}}

\renewcommand{\arraystretch}{1.2}

\begin{document}
\setconftopline{2025 IEEE Smart World Congress (SWC)}
\title{Small Language Models for Emergency Departments Decision Support: A Benchmark Study}

\author{
  \IEEEauthorblockN{
    Zirui Wang\IEEEauthorrefmark{1},
    Jiajun Wu\IEEEauthorrefmark{1},
    Braden Teitge\IEEEauthorrefmark{2},
    Jessalyn Holodinsky\IEEEauthorrefmark{3},
    Steve Drew\IEEEauthorrefmark{1}
  }
  \IEEEauthorblockA{\IEEEauthorrefmark{1}Department of Electrical and Software Engineering, University of Calgary, Calgary, AB, Canada\\
  \IEEEauthorblockA{\IEEEauthorrefmark{3}Department of Emergency Medicine, University of Calgary, Calgary, AB, Canada}  \IEEEauthorblockA{\IEEEauthorrefmark{2}Rockview General Hospital, Calgary, AB, Canada}
  \{zirui.wang1, jiajun.wu1, jessalyn.holodinsky, steve.drew\}@ucalgary.ca,  braden.teitge@albertahealthservices.ca}
}

\maketitle

\begin{abstract}
Large language models (LLMs) have become increasingly popular in medical domains to assist physicians with a variety of clinical and operational tasks. Given the fast-paced and high-stakes environment of emergency departments (EDs), small language models (SLMs), characterized by a reduction in parameter count compared to LLMs, offer significant potential due to their inherent reasoning capability and efficient performance. This enables SLMs to support physicians by providing timely and accurate information synthesis, thereby improving clinical decision-making and workflow efficiency. In this paper, we present a comprehensive benchmark designed to identify SLMs suited for ED decision support, taking into account both specialized medical expertise and broad general problem-solving capabilities. In our evaluations, we focus on SLMs that have been trained on a mixture of general-domain and medical corpora. A key motivation for emphasizing SLMs is the practical hardware limitations, operational cost constraints, and privacy concerns in the typical real-world deployments. Our benchmark datasets include MedMCQA, MedQA-4Options, and PubMedQA, with the medical abstracts dataset emulating tasks aligned with real ED physicians' daily tasks. Experimental results reveal that general-domain SLMs surprisingly outperform their medically fine-tuned counterparts across these diverse benchmarks for ED. This indicates that for ED, specialized medical fine-tuning of the model may not be required.
\end{abstract}

\begin{IEEEkeywords}
Small language models (SLMs), Emergency department (ED), Benchmarking
\end{IEEEkeywords}

\section{Introduction}

In recent years, Canadian EDs have faced significant challenges, including a significant increase in patient volume, limited resources, and longer wait times \cite{varner2023emergency, cihi2023commonwealth}.
In particular, Artificial Intelligence (AI) in the form of LLMs has tremendous potential for addressing those issues. By assisting clinicians with diagnostic reasoning, generating handoff notes, and assessing patients' acuity, AI can potentially reduce the burden on ED personnel \cite{williams2024use, hartman2024developing, goh2024large}. However, because advanced LLMs require extensive computing resources and specialized hardware infrastructure, they are not suitable for deployment in most hospital settings \cite{dennstadt2025implementing}. Furthermore, the Canadian government enacts stricter privacy regulations than those of the United States under the Health Insurance Portability and Accountability Act (HIPAA). Consequently, these regulations complicate the integration of third-party, cloud-based AI tools, such as GPT, since such tools often require routing sensitive information through external servers, which may be located outside of the country \cite{ahs2023directive}. In response to these concerns, Alberta Health Services (AHS) in Canada issued a directive on August 25, 2023, instructing its employees not to load any confidential AHS health, personal, or business information into unapproved AI programs such as ChatGPT~\cite{canadaDataResidency}.

Compared to that, Small Language Models (SLMs) have seen rapid adoption and growth in the medical domain because they're practical, cost-effective, and feasible to deploy locally \cite{garg2025rise}. Unlike their larger counterparts, SLMs can be fine-tuned efficiently for specific clinical tasks \cite{anisuzzaman2024fine, labrak2023zero}, making them particularly suitable for resource-constrained environments like the ED. As an added benefit, the use of SLMs minimizes the need to use external APIs, which ensures that sensitive patient information remains on-site, thus enhancing privacy and complying with stringent data protection regulations \cite{wang2024comprehensive}. SLMs are also capable of achieving clinically relevant performance without the computational overhead associated with their larger counterparts \cite{kim2025small}. Specifically, their reduced complexity makes them suitable for running directly on edge devices, such as NVIDIA Jetson Orin Nano \cite{dhar2024empirical, wang2025empowering}. 

We select SLMs in the range of 6 billion to 8 billion parameters that have been trained or fine-tuned on different combinations of general and medical corpora \cite{meta_llama3_8b, eleutherai_gptj6b, qwen2_7b, phi3_mini_128k, phi3_small_8k, mistral_7b_v0_3, llama3_chatqa_8b, glm_4_9b_chat, gemma_2_9b, medical_llama3_8b, llama3_8b_medqa, biogpt_large, phi3_medical_instruct, falconsai_medical_sum, t5_medical_simplifier, autotraub_med_sum, bert_medical_ner}. We include models that were originally trained on general text and then fine-tuned on medical data, as well as models primarily trained on biomedical text and others trained on a combination of both. This selection enables us to investigate how the composition of the training corpus affects performance on our ED-oriented tasks. By benchmarking these varied SLMs, we can identify which training strategy, general vs. medical, yields the best overall performance for supporting ED physicians.

To systematically explore this balance, we propose a tailored benchmark suite to evaluate language models for supporting ED physicians. Our benchmarks are designed to reflect realistic tasks and information needs that arise in the ED. In particular, we select four datasets that cover a spectrum of relevant challenges: 1. MedMCQA~\cite{medmcqa2022} for testing core medical knowledge across various subjects; 2. MedQA-4Options~\cite{medqa2020} for testing clinical problem-solving ability; 3. PubMedQA~\cite{pubmedqa2019} for testing understanding in medical literature; and 4. The Medical Abstracts dataset~\cite{medicalabstracts2020} for evaluating the model's ability to interpret medical information by simulating the process of rapidly reading relevant literature during a case. These benchmarks were chosen to closely align with real ED decision-making scenarios, ranging from recalling established medical knowledge and protocols (MedMCQA, MedQA) to retrieving and reasoning over medical research information on the fly (PubMedQA, Medical Abstracts). By covering both recall of prior knowledge and integration of new evidence, the suite comprehensively assesses a model’s fitness to assist in emergency care situations.

    Our experiments produce an intriguing conclusion: general-domain small LLMs outperform their medical-domain counterparts across all four benchmark tasks. In other words, a model with broad knowledge from diverse sources tended to answer medical questions more accurately than a model of similar size that had been fine-tuned exclusively on medical text. This finding is consistent with recent observations that expansive pre-training can endow models with strong reasoning abilities and a surprisingly robust grasp of specialized subjects. It suggests that, for the current generation of SLMs, the benefits of extensive general knowledge and reasoning skills outweigh the gains from narrow medical specialization on these ED-relevant tasks. These insights have important implications for the future development of clinical AI systems: they hint that a generalist SLM may be more reliable as a first-line ED assistant, and that adding medical training data should be done carefully to avoid over-narrowing the model’s scope. In summary, our work makes the following contributions:
\begin{itemize}
    \item \textbf{Benchmark Suite for ED-Focused SLM Evaluation:} We collected a set of benchmark tasks that reflect real ED workflows, providing a testbed for SLMs intended to support ED physicians. Our suite covers both medical knowledge recall and general reasoning, enabling a balanced assessment of model capabilities in this domain.
    \item \textbf{Evaluation of SLMs:} We present a thorough evaluation of SLMs that are feasible to deploy under typical ED hardware and cost constraints. To our knowledge, this is one of the first studies to emphasize SLMs in ED, highlighting their potential for practical, on-premise use.
    \item \textbf{Insights on SLMs Design for ED:} Our findings show that models with general pre-training can outperform domain-specialized models on ED tasks, suggesting that diverse knowledge sources might contribute critically to performance in emergency medicine scenarios. Our study challenges the assumption that fine-tuning medical AI in a particular domain always results in better performance, instead suggesting that maintaining broad-ranging reasoning skills might prove beneficial in the context of emergency medicine. Lastly, we discuss the implications of this result for the development of future AI systems for the ED.
\end{itemize}

\section{Related Works}
\subsection{General Benchmarks}

General-purpose LLMs, such as OpenAI's GPT-4 and Meta’s Llama series, are trained on vast and diverse datasets covering a wide range of topics. This broad training allows these models to develop strong general knowledge and robust language understanding skills~\cite{brown2020language, openai2023gpt4, touvron2023llama}. Thanks to this versatility, general-purpose models can adapt to various tasks, even those that traditionally require domain-specific expertise, such as standardized medical exams. Recent studies have shown that these models can sometimes match or even surpass human-level accuracy in medical question-answering tasks, even without any medical-specific fine-tuning~\cite{kung2023performance, nori2023capabilities}.

\subsection{Medical-Specific Benchmarks}

On the other hand, medical-domain-specific LLMs are fine-tuned on curated medical datasets to improve their performance in specialized tasks. Notable examples include Google’s Med-PaLM 2, Microsoft’s BioGPT, and open-source models like MedAlpaca, Llama 3-Medical, and Qwen-Medical~\cite{singhal2023expertlevel}. These models have demonstrated strong results in areas such as medical question answering, clinical text summarization, and understanding biomedical literature~\cite{singhal2023expertlevel}. However, despite their theoretical advantages, access to many of these specialized models remains limited. Moreover, there is still a lack of conclusive, real-world evidence showing that they consistently outperform general-purpose models in day-to-day clinical practice~\cite{lee2023benefits}.

\subsection{Resource and Deployment-Focused Benchmarks}

Much of the current benchmarking work focuses on general medical exams or synthetic datasets, which often fail to reflect the realities of clinical workflows. In particular, there is a notable lack of benchmarks that are specifically tailored to the unique challenges of Canadian EDs. These settings demand fast decision-making, handle high patient volumes, require efficient information management, and often face significant resource constraints~\cite{chan2023emergency, mahajan2022operational}. These operational factors have a major impact on the practical utility of AI tools, but they are rarely considered in standard evaluations.

Crucially, no existing benchmarks explicitly target ED scenarios that combine realistic clinical tasks with the severe resource limitations that are common in these environments. This gap makes it hard to evaluate whether current LLMs are truly ready for deployment in the ED.

To address this issue, our study benchmarks both general-purpose and medical-specific LLMs on datasets that were carefully chosen to reflect real-world ED tasks and workflows—MedMCQA, MedQA 4-options, PubMedQA, and Medical Abstracts~\cite{pal2022medmcqa, jin2021pubmedqa}. By evaluating models in a more realistic, operationally focused context, we aim to identify the LLMs that are best suited for practical use in Canadian EDs. Our goal is to help healthcare institutions adopt AI tools that are specifically designed to tackle critical challenges in the ED, such as physician burnout, long patient wait times, and inefficiencies in clinical workflows~\cite{west2022clinician, lane2023waittimes}.

\section{Methodology}
\subsection{Model Selection}
We began by screening approximately 50 open-source LLMs to assess their suitability for deployment in real-world, resource-limited hospital settings, specifically Canadian EDs. In selecting these models, we prioritized several key factors that directly impact deployment feasibility and clinical usefulness. First, we focused on transformer-based architectures that offer modern, efficient design principles optimized for scalability and speed. Given the hardware constraints typically present in hospitals, we limited our selection to models with fewer than 8 billion parameters to ensure they can run on commonly available GPUs without requiring high-end infrastructure. Another important consideration was the diversity and relevance of the training data; we looked for models trained on a broad range of text, including general language corpora and, where possible, medical datasets, to ensure robust understanding across both clinical and conversational contexts. Finally, we took into account the reported performance of models in existing natural language processing (NLP) benchmarks, particularly in tasks like question answering and summarization, as these reflect the types of cognitive challenges that physicians frequently encounter in the ED. After applying these criteria, we identified 17 models for detailed evaluation—nine general-purpose models and eight medical-specialized models. All of these models have parameter counts of 8 billion or fewer, ensuring they remain practical for deployment in hospital settings where hardware resources are limited. For general-purpose models, these models were selected for their broad language understanding capabilities, scalability, and potential to support a wide range of clinical tasks.

\begin{table*}[htbp]
  \caption{Overview of General and Medical–Specialized Models for ED Deployment}
  \label{tab:ed_model_overview}
  \centering
  \normalsize
  \begin{tabularx}{\linewidth}{@{}p{5.5cm}X@{}}
    \toprule
    \multicolumn{2}{l}{\textbf{General Models}} \\
    \midrule
    \textbf{Meta Llama-3-8B}~\cite{meta_llama3_8b} & A state-of-the-art general model with strong zero-shot reasoning performance. Its open-access availability aligns with Canadian ED privacy requirements, and its moderate size allows deployment on hospital hardware while offering robust diagnostic support. \\
    \textbf{EleutherAI GPT-J-6B}~\cite{eleutherai_gptj6b} & A classic open-source model serving as a baseline for general language reasoning. Though not tailored for medical tasks, its inclusion highlights the gap between purely general models and those adapted for healthcare. \\
    \textbf{Qwen2-7B}~\cite{qwen2_7b} & Chosen for its strong multilingual and instruction-following capabilities, key for Canadian EDs serving diverse patient populations. \\
    \textbf{Microsoft Phi3-mini-128k}~\cite{phi3_mini_128k} & A compact, instruction-tuned model designed for efficiency, making it suitable for tasks like triage note summarization in EDs with limited hardware. \\
    \textbf{Microsoft Phi3-small-8k}~\cite{phi3_small_8k} & Balances performance and size, ideal for tasks requiring quick turnaround, such as discharge summaries or guideline lookups. \\
    \textbf{Mistral-7B-v0.3}~\cite{mistral_7b_v0_3} & A cutting-edge model with strong reasoning abilities, selected to evaluate how well general models handle complex ED tasks like differential diagnosis. \\
    \textbf{Llama3-ChatQA-8B}~\cite{llama3_chatqa_8b} & Fine-tuned for conversational tasks, supporting interactive diagnostic Q\&A and bedside guideline explanations. \\
    \textbf{GTHUDM GLM-4-9B-chat}~\cite{glm_4_9b_chat} & A slightly larger model included to explore the trade-off between resource demands and performance on reasoning-heavy tasks. \\
    \textbf{Google Gemma-2-9B}~\cite{gemma_2_9b} & A compact, efficient model evaluated for its general reasoning abilities, included as an open-weight option for academic research in Canadian EDs. \\
    \midrule
    \multicolumn{2}{l}{\textbf{Medical–Specialized Models}} \\
    \midrule
    \textbf{Medical-Llama-3-8B}~\cite{medical_llama3_8b} & A medical variant of Llama-3, focused on clinical reasoning and question answering, selected for its balance of size and task relevance. \\
    \textbf{Llama-3-8B-MedQA}~\cite{llama3_8b_medqa} & An instruction-tuned model designed for fast, accurate clinical question answering—critical in ED settings. \\
    \textbf{Microsoft BioGPT-Large}~\cite{biogpt_large} & Pretrained on biomedical texts, evaluated for its ability to support literature summarization and evidence retrieval. \\ 
    \textbf{Phi-3-medical-instruct}~\cite{phi3_medical_instruct} & Fine-tuned for medical instruction tasks, relevant for guideline summarization and treatment explanations. \\
    \textbf{FalconSaI-medical-sum(T5-based)}~\cite{falconsai_medical_sum} & Focused on summarizing complex medical documents vital for condensing lengthy patient histories into actionable summaries. \\
    \textbf{T5-medical-simplifier}~\cite{t5_medical_simplifier} & A model designed to simplify medical text, supporting clear communication in diverse patient populations. \\
    \textbf{AutoTraub-med-sum}~\cite{autotraub_med_sum} & Targets medical report summarization, helping ED physicians manage multiple patient charts efficiently. \\
    \textbf{BERT-medical-NER}~\cite{bert_medical_ner} & Primarily for Chinese medical entity recognition, included to explore multilingual support in Canadian EDs. \\
    \bottomrule
  \end{tabularx}
\end{table*}

\vspace{0.5em}

\subsection{Benchmarking Tasks and Datasets}

We benchmarked the models on four datasets, each representing a core information-processing need in Canadian EDs. MedMCQA (MCQA)~\cite{medmcqa2022} tests a model’s breadth of medical knowledge and its ability to reason through clinical facts and exam-style questions. MedQA (USMLE 4-Options) (MQ4)~\cite{medqa2020} focuses on clinical problem-solving, presenting realistic patient scenarios with multiple plausible answers. PubMedQA (PMQA)~\cite{pubmedqa2019} assesses the model’s ability to extract, interpret, and apply evidence from biomedical research abstracts. Medical Abstracts (MA)~\cite{medicalabstracts2020} evaluates how well the model can condense dense research abstracts into concise, clinically useful summaries.

These tasks were selected to reflect the diverse challenges physicians face in Canadian EDs, from rapid clinical decision-making to accurate documentation and evidence-based support.

\subsection{Experimental Setup}

All evaluations were conducted using the standardized \texttt{lm-evaluation-harness} framework~\cite{lm_eval_harness} on Google Colab Pro, utilizing a single NVIDIA A100 GPU with 40GB of VRAM.


\paragraph{\textbf{Model Quantization}}  
To ensure realistic deployment feasibility for ED settings, we applied model quantization (4-bit and 8-bit) using Hugging Face’s \texttt{transformers} library integrated with \texttt{bitsandbytes}. This significantly reduced memory usage while maintaining model accuracy.


\paragraph{\textbf{Batch Sizes}}  
Batch sizes were carefully selected to make the most of available GPU resources while avoiding memory overflows. Smaller models, defined as those with fewer than 8 billion parameters, were evaluated with a batch size of 8. For larger models, around 8 billion parameters, we used a batch size of 4 to accommodate increased memory demands.

\subsection{Evaluation Metrics}



For the multiple-choice question answering tasks: MedMCQA, MedQA (4-options), and PubMedQA, we report accuracy as the primary metric. For the medical abstract summarization task, we evaluate model performance using three metrics from the Unitxt framework: Unitxt Accuracy (MA), Macro-F1 (F1-Ma), and Micro-F1 (F1-Mi). Hence, these metrics provide a comprehensive assessment of summary quality, capturing not only correctness but also the balance between precision and recall, as well as the degree of semantic overlap with reference texts.

\paragraph{Accuracy (ACC)}  
Accuracy measures how often the model selects the correct option in multiple-choice tasks. It is a straightforward indicator of correctness, especially relevant when there is a single correct answer. Formally, we calculate accuracy as:
\[
  \mathrm{ACC} = \frac{1}{N} \sum_{i=1}^N \mathbf{1}\{\hat{y}_i = y_i\},
\]
where $\hat{y}_i$ is the model’s prediction, and $y_i$ is the correct answer.


\paragraph{UniTxt Accuracy}  
For summarization tasks, UniTxt Accuracy evaluates how often the model’s output exactly matches the reference summary after normalization. This metric provides insight into the model’s ability to generate precise, high-fidelity summaries, which is critical in clinical documentation. The formula for UniTxt Accuracy is:
\[
  \mathrm{UniTxtAccuracy} = \frac{\#\{\text{exact matches}\}}{N}.
\]


\paragraph{F1 Score}  
The F1 Score helps assess how much of the important information the model captures by combining precision (the proportion of correct tokens in the output) and recall (the proportion of relevant tokens the model correctly identifies). It is calculated as:
\[
  \mathrm{Prec} = \frac{\mathrm{TP}}{\mathrm{TP}+\mathrm{FP}}, \quad
  \mathrm{Rec}  = \frac{\mathrm{TP}}{\mathrm{TP}+\mathrm{FN}},
\]
and
\[
  F1 = 2 \cdot \frac{\mathrm{Prec} \cdot \mathrm{Rec}}{\mathrm{Prec} + \mathrm{Rec}}.
\]
We report two types of F1 scores in this study: Macro-F1, which is the average of the F1 scores across all individual examples and treats each case equally, providing a case-by-case perspective; and Micro-F1, which is a global F1 score computed by pooling all true positives, false positives, and false negatives, offering an overall sense of the model’s performance across the entire dataset.


\paragraph{Why These Metrics?}  
Each metric offers a unique perspective on model performance. Accuracy indicates how often the model produces the exact correct answer, a critical aspect for multiple-choice medical question tasks. The F1 Score captures how well the model balances precision and recall, which is essential for complex tasks like summarization and detailed question answering. UniTxt Accuracy, on the other hand, measures whether the model can exactly reproduce the reference answer, an important consideration in clinical settings where even small errors can have significant consequences.


\section{{Experimental Results and Discussion}}
We evaluated both general-purpose and medical-specialized language models across four clinically relevant tasks designed to simulate real-world ED scenarios. These tasks include multiple-choice question answering using the MedMCQA and MedQA (4-option) datasets, evidence-based reasoning via PubMedQA, and summarization of medical abstracts. Together, they reflect key applications such as diagnostic support, clinical decision-making, and summarizing patient records or literature. For brevity in the tables, we use the following abbreviations: \texttt{MCQA} (MedMCQA), \texttt{MQ4} (MedQA 4‑Options), \texttt{PMQA} (PubMedQA), \texttt{MA} (Medical Abstracts), \texttt{F1-Ma} (F1 Macro), and \texttt{F1-Mi} (F1 Micro). Performance results for all models are presented in Tables~\ref{tab:general_llms} and~\ref{tab:medical_llms}.

Before presenting the benchmarking results, we clarify that several models have missing values for the Medical Abstract (MA) task and its associated F1 metrics. In Tables~\ref{tab:general_llms} and~\ref{tab:medical_llms}, these missing entries are denoted by "---", and are not caused by evaluation errors. Rather, they reflect architectural limitations or task-model mismatches. For example, Microsoft Phi3-small-8k and Microsoft Phi3-mini-128k are general-purpose models not fine-tuned for summarization tasks. Phi-3-medical-instruct is optimized for medical question answering but not for structured summarization. Similarly, T5-medical-simplifier and BERT-medical-NER are designed for sentence simplification and named entity recognition, respectively, and do not apply to summarization or multiple-choice QA tasks.

\begin{table}[htbp]
  \centering
  \caption{Benchmark Scores for General-Purpose SLMs}
  \label{tab:general_llms}
  \resizebox{1\columnwidth}{!}{%
  \begin{tabular}{l|cccccc}
    \toprule
    \textbf{Model} & \textbf{MCQA} & \textbf{MQ4} & \textbf{PMQA} & \textbf{MA} & \textbf{F1-Ma} & \textbf{F1-Mi} \\
    \midrule
    Meta Llama-3-8B       & 0.5750 & 0.5970 & 0.7480 & 0.2432 & 0.2431 & 0.1415 \\
    EleutherAI GPT-J-6B   & 0.2986 & 0.2569 & 0.5780 & 0.1579 & 0.0726 & 0.1821 \\
    Qwen2-7B              & 0.5699 & 0.5672 & 0.7410 & 0.3317 & 0.2239 & \textbf{0.3502} \\
    Microsoft Phi3-mini-128k& 0.5243 & 0.5357 & 0.7520 & ---  & ---  & --- \\
    Microsoft Phi3-small-8k& \textbf{0.6036} & \textbf{0.6386} & \textbf{0.7760} & --- & --- & --- \\
    Mistral-7B-v0.3       & 0.4800 & 0.5027 & 0.7560 & 0.2195 & 0.2192 & 0.0721 \\
    Llama3-ChatQA-8B      & 0.5264 & 0.5530 & 0.7160 & 0.3149 & \textbf{0.3144} & 0.2929 \\
    GTHUDM GLM-4-9B-chat  & 0.5183 & 0.6096 & 0.7380 & \textbf{0.3369} & 0.2650 & 0.3370 \\
    Google Gemma-2-9B     & 0.2670 & 0.2820 & 0.5560 & 0.0003 & 0.0004 & 0.0007 \\
    \bottomrule
  \end{tabular}%
  }
\end{table}

\begin{table}[htbp]
  \centering
  \caption{Benchmark Scores for Medical-Specialized SLMs}
  \label{tab:medical_llms}
  \resizebox{1\columnwidth}{!}{%
  \begin{tabular}{l|cccccc}
    \toprule
    \textbf{Model} & \textbf{MCQA} & \textbf{MQ4} & \textbf{PMQA} & \textbf{MA} & \textbf{F1-Ma} & \textbf{F1-Mi} \\
    \midrule
    Medical-Llama-3-8B       & \textbf{0.5368} & \textbf{0.5719} & 0.7400 & 0.2458 & 0.1432 & 0.2459 \\
    Llama-3-8B-MedQA         & 0.4939 & 0.5460 & 0.7000 & \textbf{0.3199} & \textbf{0.2594} & \textbf{0.3447} \\
    Microsoft BioGPT-Large   & 0.3199 & 0.2836 & 0.6120 & 0.0215 & 0.0291 & 0.0399 \\
    Phi-3-medical-instruct   & 0.5173 & 0.5200 & \textbf{0.7520} & --- & --- & ---  \\
    FalconSaI-medical-sum(T5-based)& 0.3175 & 0.2757 & 0.3480 & 0.0017 & 0.0029 & 0.0034 \\
    T5-medical-simplifier    & 0.3215 & 0.2773 & 0.3640 & --- & --- & --- \\
    AutoTraub-med-sum        & 0.2462 & 0.2514 & 0.4280 & 0.2050 & 0.1985 & 0.2946 \\
    BERT-medical-NER         & 0.3223 & 0.2773 & 0.5520 & --- & --- & ---  \\
    \bottomrule
  \end{tabular}%
  }
\end{table}
Note: "---" indicates the model does not apply to the task. See the above text for details. MCQA, MQ4, and PMQA are evaluated using accuracy. MA is evaluated using Unitxt Accuracy, Macro-F1 (F1-Ma), and Micro-F1 (F1-Mi).

\subsection{Q\&A Performance (MCQA and MQ4)} 

The instruction-tuned Microsoft \texttt{Phi3-small-8k} model demonstrated exceptional performance across all QA benchmarks, outperforming even specialized medical models. Its strong zero-shot reasoning capabilities reflect the benefits of advanced instruction tuning combined with a lightweight architecture. In contrast, general-purpose models like \texttt{EleutherAI GPT-J-6B} and \texttt{Google Gemma-2-9B} consistently underperformed, indicating limited domain adaptability for medical QA tasks.


\subsection{Domain Knowledge vs. Instruction Tuning (PMQA)}
Microsoft \texttt{Phi3-small-8k} achieved higher accuracy on PubMedQA than \texttt{Medical-Llama-3-8B}, despite lacking medical-specific fine-tuning. This suggests that strong instruction tuning on broad datasets can sometimes outperform domain-focused models in multi-step reasoning tasks. However, models like \texttt{Phi-3-medical-instruct} still led in evidence-based QA (75.2\% on PubMedQA) in medical models, reinforcing the value of domain specialization for certain clinical tasks.


\subsection{Summarization Results (MA)}  
Summarization of medical abstracts proved challenging. Only chat-optimized models like \texttt{GTHUDM GLM-4-9B-chat} and \texttt{Llama3-ChatQA-8B} generated reasonably accurate summaries, with F1 scores above 0.29. Models not specifically tuned for summarization, such as \texttt{Phi3-small series} and \texttt{T5-medical-simplifier}, consistently returned null or irrelevant outputs. This underscores the limitations of applying QA-tuned models to generation tasks without dedicated adaptation.


\subsection{Specific Observations and Key Takeaways}  
The detailed evaluation of both general-purpose and medical-specialized models revealed several important trends.  
Notably, general-purpose models demonstrated strong and often surprising capabilities across diverse tasks.  
For example, \texttt{Qwen2-7B} showed balanced performance, achieving the highest micro-F1 score (0.3502) and consistent token-level precision.  
\texttt{AutoTraub-med-sum} exhibited potential in summarization (F1-Mi = 0.2946) but struggled in QA settings, suggesting it may be more suited for targeted applications such as report generation.  
\texttt{Medical-Llama-3-8B} performed well in QA and biomedical comprehension tasks but showed limited ability in summarization, indicating that domain specialization alone may not be sufficient for generative tasks.  
Conversely, models such as \texttt{FalconSal-medical-sum(T5-based)}, \texttt{Microsoft BioGPT-Large}, and \texttt{Gemma-2-9B} consistently underperformed across all tasks, highlighting the need for further tuning and adaptation before deployment in ED scenarios.

Overall, our findings suggest that \texttt{Microsoft Phi3-small-8k} provides the best balance of performance and deployability for QA tasks in ED, while \texttt{GTHUDM GLM-4-9B-chat} and \texttt{Llama3-ChatQA-8B} are preferred options for summarization, particularly for condensing patient histories and medical literature.  
Although domain-specific models like \texttt{Medical-Llama-3-8B} remain strong candidates for inference-focused scenarios, the results indicate that a hybrid approach—combining QA-specialized and summarization-specialized models may offer a more practical and effective solution for real-world ED deployments.


\subsection{General Models vs. Medical Models}
Across our benchmarks, we consistently observed that general-purpose models tended to outperform medical-specialized models on multiple tasks, particularly in question-answering scenarios. This is a noteworthy and encouraging signal. It suggests that well-tuned general models, when designed with strong instruction-following capabilities and broad linguistic knowledge, can transfer effectively to medical domains without explicit medical fine-tuning. This has significant implications: it indicates that hospitals and research teams may not always need access to proprietary, domain-specific models to achieve meaningful results in certain ED tasks. Instead, with careful selection and deployment, open-access general models could offer a practical, cost-effective alternative for institutions with limited resources, reducing barriers to adopting AI tools in healthcare.

\subsection{Future Work}
Our benchmarking results offer a useful snapshot of how current SLMs perform in ED settings. However, closing the gap between these experimental outcomes and practical clinical adoption will require substantial future work. Several key directions stand out as especially important for enabling real-world impact in Canadian EDs.

One foundational area is domain-specific fine-tuning. Off-the-shelf SLMs still lack sensitivity to the nuances of ED language, urgency, and workflows. A phrase like “abdominal pain with guarding” implies very different clinical concerns than “abdominal pain, non-specific,” but a general-purpose model might treat both similarly. Fine-tuning on targeted datasets, triage assessments, shift handovers, resuscitation notes, and discharge summaries can help models grasp the shorthand, priorities, and high-stakes reasoning embedded in ED documentation. To be widely applicable, this training should draw from a diverse corpus reflecting urban and rural hospitals, and ideally support both English and French.

Beyond fine-tuning, there’s a strong case for developing models tailored to specific high-value ED tasks. These include triage support (e.g., recommending Canadian Triage and Acuity Scale levels), differential diagnosis generation for common complaints like chest pain or shortness of breath, and drafting discharge instructions that are clear, accurate, and accessible to patients. Handing off care between teams is another pain point in EDs; language models could help produce structured, SBAR-style summaries that ensure nothing critical is lost in transition. Each of these use cases likely demands its own prompt format, training regimen, and evaluation criteria.


Infrastructure is just as important as intelligence. Many EDs in Canada, particularly in rural or northern regions, cannot depend on stable cloud access. Offline-capable deployment pipelines running on hospital-grade GPUs like the NVIDIA L4 or A100, or even on CPUs using quantized models, will be essential. Techniques like LoRA (Low-Rank Adaptation), quantization-aware training, and sparsity optimizations could allow high-performance models to run in under 500ms per query without massive compute.

No single model will be able to do everything well. A modular, multi-agent approach might be more practical, where specialized SLMs handle subtasks such as summarizing clinical notes, retrieving medical guidelines, or answering physician queries. An orchestration layer could route requests intelligently, much like real ED teams delegate and collaborate. Designing and testing such architectures remains an open challenge.

Model transparency and trust are also critical. ED physicians need to know why a model recommends a certain action, not just what it recommends. Ideally, outputs should be accompanied by justifications like: “CTAS Level 2 recommended due to hypotension and chest pain with diaphoresis, based on HEART score criteria.” Uncertainty estimates and references to clinical guidelines (e.g., UpToDate or CAEP policies) can further support trust and clinical defensibility.

Lastly, all of this work must be tested in real-world settings. Controlled pilots in Canadian EDs, starting with lower-risk applications like triage note summarization or discharge instruction drafting, can help assess feasibility, utility, and risks. A trial in a rural Alberta ED, for example, could evaluate whether a model improves documentation completeness and clinician satisfaction during off-hours. Key metrics might include time saved per patient, physician feedback, and error rates. These pilots would also surface integration barriers such as alert fatigue, workflow disruption, and the limits of current human-AI collaboration.

Taken together, these directions outline a path not just for building better models, but for designing systems that work in the complex, time-pressured, and high-stakes environment of emergency care.

\section{Practical Recommendations}

To enable safe, effective, and contextually relevant deployment of SLMs in Canadian EDs, we provide a set of recommendations grounded in our benchmarking results and a realistic understanding of hospital operational constraints.

\begin{table}[htbp]
\scriptsize
\centering
\caption{SLMs Suited for Use in Canadian EDs}
\label{tab:recommended_slms}
\renewcommand{\arraystretch}{1.15}
\begin{tabular}{|>{\raggedright\arraybackslash}p{1.7cm}|>{\raggedright\arraybackslash}p{3.0cm}|>{\raggedright\arraybackslash}p{2.0cm}|}
\hline
\textbf{Model} & \textbf{Use Case(s)} & \textbf{Hardware Fit} \\
\hline
\textbf{Phi3-small-8k} & Question answering, differential diagnosis generation, guideline lookup & RTX 4090 / L4; real-time QA with 4–8-bit quant. \\
\hline
\textbf{Qwen2-7B} & Balanced QA and summarization; multilingual support for diverse patient populations & A100 / L4 / T4; general ED use \\
\hline
\textbf{GLM-4-9B-chat} & Medical abstract/history summarization; conversational agent & A100 (40GB)+; long-input tasks \\
\hline
\end{tabular}
\end{table}

Effective deployment in EDs must also account for hardware limitations, latency requirements, and task specialization. All models should undergo 4 or 8 bit quantization using frameworks such as \texttt{bitsandbytes} or \texttt{GPTQ}. Quantization reduces memory usage by 50–75\%, allowing real-time inference on mid-range GPUs like the RTX 4090 or A100 with minimal performance degradation. Our benchmarks show that \textbf{Microsoft Phi3-small-8k} maintained near state-of-the-art QA performance post-quantization. Fine-tuning on anonymized Canadian ED datasets, including triage assessments, discharge notes, and critical care summaries, will ensure models can handle ED-specific language (e.g., “CTAS Level 2,” “R/O STEMI”) and generate realistic outputs, such as SOAP notes or risk stratifications like HEART scores for chest pain. Inference optimization with frameworks like \texttt{ONNX Runtime}, \texttt{vLLM}, or \texttt{TensorRT} is recommended to reduce latency, which is critical for tasks such as generating a sepsis alert within 500 milliseconds. For hardware, a single-node GPU system such as an RTX 4090 or NVIDIA L4 will meet most ED needs for real-time inference with quantized models. Larger hospitals may benefit from edge clusters to support multi-user access during peak hours.

Operational considerations are critical when deploying SLMs in Canadian EDs. All deployments must comply with Canadian privacy laws, such as the Health Information Act in Alberta, and relevant data governance standards. Inference and data processing should occur within the hospital’s secure network, as reliance on external APIs such as OpenAI’s GPT-4, hosted in the U.S., risks non-compliance with regulations like PHIPA. Models must also support offline-first deployment to ensure reliability in rural EDs where network connectivity is inconsistent. Every output should be logged with timestamps, input queries, and, when feasible, model explanations, for example: “Suggested CTAS Level 2 based on chest pain, diaphoresis, and hypotension, per HEART guidelines.” Human-in-the-loop supervision remains essential, as licensed healthcare professionals must review and approve AI-generated outputs. For instance, while a model may suggest a differential diagnosis such as “pulmonary embolism versus pneumonia,” the final decision must rest with the attending physician.

\section{Conclusion}
This paper introduced a comprehensive benchmark evaluation of SLMs for supporting physician decision-making in the ED. We evaluated various SLMs using multiple ED-relevant benchmarks, such as MedMCQA, MedQA-4Options, PubMedQA, and the Medical Abstracts dataset. Our findings revealed that general-domain SLMs outperformed models fine-tuned or pre-trained on medical corpora. This suggests that, within the context of emergency medicine, robust general reasoning and broad knowledge bases may be more critical than specialized medical training. It is our hope that the findings of our work will have meaningful implications for the development of SLM in the ED. Practitioners and researchers should pay more attention to the general reasoning capabilities when fine-tuning SLMs for clinical use, as excessive specialization could inadvertently narrow the scope of the model's applicability. Furthermore, our proposed benchmark suite offers a valuable evaluation framework, guiding the design and selection of clinically relevant, resource-efficient SLMs suitable for deployment in the ED.

\bibliographystyle{IEEEtran}
\bibliography{references}
\end{document}